\documentclass{conm-p-l}

\usepackage{graphicx}
\usepackage{multirow}
\usepackage{algorithmicx}
\usepackage{algpseudocode}
\usepackage{algorithm}

\graphicspath{{./Images/}}

\theoremstyle{definition}

\theoremstyle{remark}

\numberwithin{equation}{section}

\begin{document}

\title[Performance of Hull-Detection Algorithms]{Performance of Hull-Detection Algorithms For Proton Computed Tomography Reconstruction}

\author[B. Schultze]{Blake Schultze}
\address[B.~Schultze]{Department of Electrical and Computer Engineering, Baylor University, Waco, TX 76798, USA }
\email[B.~Schultze]{blake{\textunderscore}schultze@baylor.edu}
\thanks{Schultze, Schubert, Schulte, and Witt were partially supported by Grant No. 1R01EB013118-01 from the National Institute of Biomedical Imaging and Bioengineering at the National Institutes of Health and the National Science Foundation. The content of this paper is solely the responsibility of the authors and does not necessarily represent the official views of the National Institute of Biomedical Imaging and Bioengineering or the National Institutes of Health. }

\author[M. Witt]{Micah Witt}
\address[M.~Witt]{Department of Computer Science and Engineering, California State University, San Bernardino, San Bernardino, CA 92407, USA}
\email[M.~Witt]{micah@r2labs.org}

\author[Y. Censor]{Yair Censor}
\address[Y.~Censor]{Department of Mathematics, University of Haifa, Haifa 3190501, Israel}
\email[Y.~Censor]{yair@math.haifa.ac.il}

\author[R. Schulte]{Reinhard Schulte}
\address[R.~Schulte]{Department of Radiation Medicine, Loma Linda University Medical Center, Loma Linda, CA 92354, USA}
\email[R.~Schulte]{rschulte@llu.edu}
\thanks{Schulte and Censor were partially supported by Grant No. 2009012 from the United States - Israel Binational Science Foundation (BSF)}

\author[K.~E. Schubert]{Keith Evan Schubert}
\address[K.~E.~Schubert]{Department of Electrical and Computer Engineering, Baylor University, Waco, TX 76798, USA}
\email[K.~E.~Schubert]{keith{\textunderscore}schubert@baylor.edu}

\subjclass{Primary 92C55, 94A08, 68U10; Secondary 65F50, 68W10}
\date{30 November 2013}


\begin{abstract}
Proton computed tomography (pCT) is a novel imaging modality developed for patients receiving proton radiation therapy. The purpose of this work was to investigate hull-detection algorithms used for preconditioning of the large and sparse linear system of equations that needs to be solved for pCT image reconstruction. The hull-detection algorithms investigated here included silhouette/space carving (SC), modified silhouette/space carving (MSC), and space modeling (SM).  Each was compared to the cone-beam version of filtered backprojection (FBP) used for hull-detection. Data for testing these algorithms included simulated data sets of a digital head phantom and an experimental data set of a pediatric head phantom obtained with a pCT scanner prototype at Loma Linda University Medical Center.  SC was the fastest algorithm, exceeding the speed of FBP by more than 100 times. FBP was most sensitive to the presence of noise. Ongoing work will focus on optimizing threshold parameters in order to define a fast and efficient method for hull-detection in pCT image reconstruction.
\end{abstract}

\maketitle
\begin{center}
December 8, 2013.  Revised: February 2, 2014.
\end{center}
\section{Introduction}
\label{sec:Introduction}
Proton computed tomography (pCT) has the potential to become a preferred imaging modality for the planning of proton radiation therapy because images reconstructed from pCT data predict the range of proton beams in the patient more accurately than those obtained from x-ray CT data~\cite{Smith09}.

The data produced by a pCT scanner are energy measurements of individual protons traversing the object from many different directions. These energy measurements are then converted to water-equivalent path length (WEPL). ``Water-equivalent'' in this context means that if the proton has the given path-length in water, it will, on average, lose the same amount of energy that the proton has lost by traversing the object. From this data, one can reconstruct the relative stopping power (RSP) of protons with respect to water by the formula $RSP=S_{mat}/S_{water}$, where the stopping power of a material, $S_{mat}=-dE/dx$, is the mean differential energy loss ($dE$) of protons per unit path length ($dx$).  The image reconstruction of pCT requires finding a computationally tractable solution to a large and sparse linear system of equations of the form $Ax=b$, where the matrix $A$ contains the proton path information, i.e., the index of the object voxels intersected by the path, the vector $b$ contains the WEPL values, and the solution vector $x$ contains the RSP values in voxels after the system has been solved.

High-speed performance and accurate reconstruction are important prerequisites for clinical applicability of pCT. The size of the linear system, which is of the order of $10^{8}\times 10^{7}$, requires a parallelizable iterative image reconstruction algorithm to be implemented across a multi-processor, e.g., a graphics processing unit (GPU), computing cluster. Recent studies demonstrated that good quality pCT images were reconstructed with iterative projection algorithms performed on a single GPU~\cite{PenfoldDis,PSCR10}.

Efficient image reconstruction can be significantly expedited if accurate knowledge of the object's hull is available.  For an object $X\subset\mathbb{N}^3$, which is a finite set of voxels, and a discrete image space $V\subset\mathbb{N}^3$ defined such that $X\subseteq V$, we define the object's hull, $H$, as the smallest subset $H\subseteq V$ such that $X\subseteq H$; in other words, the hull is the the smallest bounded region that encloses the object.  The object's hull, $H$, is needed for the most likely path (MLP) calculations, which is an important step in pCT image reconstruction \cite{SPTS08}.

With an accurate hull-detection algorithm, any voxel outside the hull can be excluded from image reconstruction, effectively reducing the size of $x$ and, thus, the number of columns of $A$.  Iterative solutions of a linear system with $m$ rows (proton histories) and $n$ columns (voxels) has time complexity $O(mnk)$, where $k$ is the number of iterations.  In a pCT system, typically $100n\ge m \ge 10n$.  However, a proton passes through the largest number of voxels when it traverses the reconstruction volume diagonally, corresponding to approximately $\sqrt[3]{n}$ voxels.  Thus, the system matrix, $A$, only contains on the order of $\sqrt[3]{n}$ non-zero elements and, consequently, $O(mnk)$ reduces to $O(n^{1.333}k)$.  The memory requirements for the parallel iterative image reconstruction algorithms also decrease as the number of voxels misidentified as part of the hull is reduced. This is particularly important in a GPU system.  An efficient hull-detection algorithm reduces the execution time of pCT image reconstruction.  Therefore, the objective of the hull-detection algorithms for pCT is to efficiently produce an accurate approximation of the hull.

In this work, we report on hull-detection using two established algorithms, i.e., filtered backprojection (FBP) and silhouette/space carving (SC) and two new algorithms, i.e., modified silhouette/spacecarving (MSC) and space modeling (SM) that we developed for pCT image reconstruction.
\section{Data Characteristics}
\label{sec:Data}
The data used in this work came from two sources.  Initial data testing was done using a simulated digital phantom designed for pCT~\cite{WSS12}.  Advanced testing was done on an actual scan of a pediatric head phantom using the Phase I prototype pCT scanner at Loma Linda University Medical Center~\cite{Fel07,Cut07}.  The simulated and scanned data sets are described in the following sections.

\subsection{Simulated Phantom}
\label{ssec:Phantom and Input Data}
The phantom input data used to assess the performance of each hull-detection algorithm was produced using a pCT simulator specifically designed for algorithm analysis~\cite{WSS12}. The simulator provides the user with the ability to construct a non-homogeneous elliptical object (NEO) to approximate head phantoms of various sizes and with simplified representations of anatomical features, such as ventricles, frontal sinus, ears, and nose.

\begin{figure}[htb]
\begin{minipage}[b]{1.0\linewidth}
  \centering
  \centerline{\includegraphics[width=4.0cm]{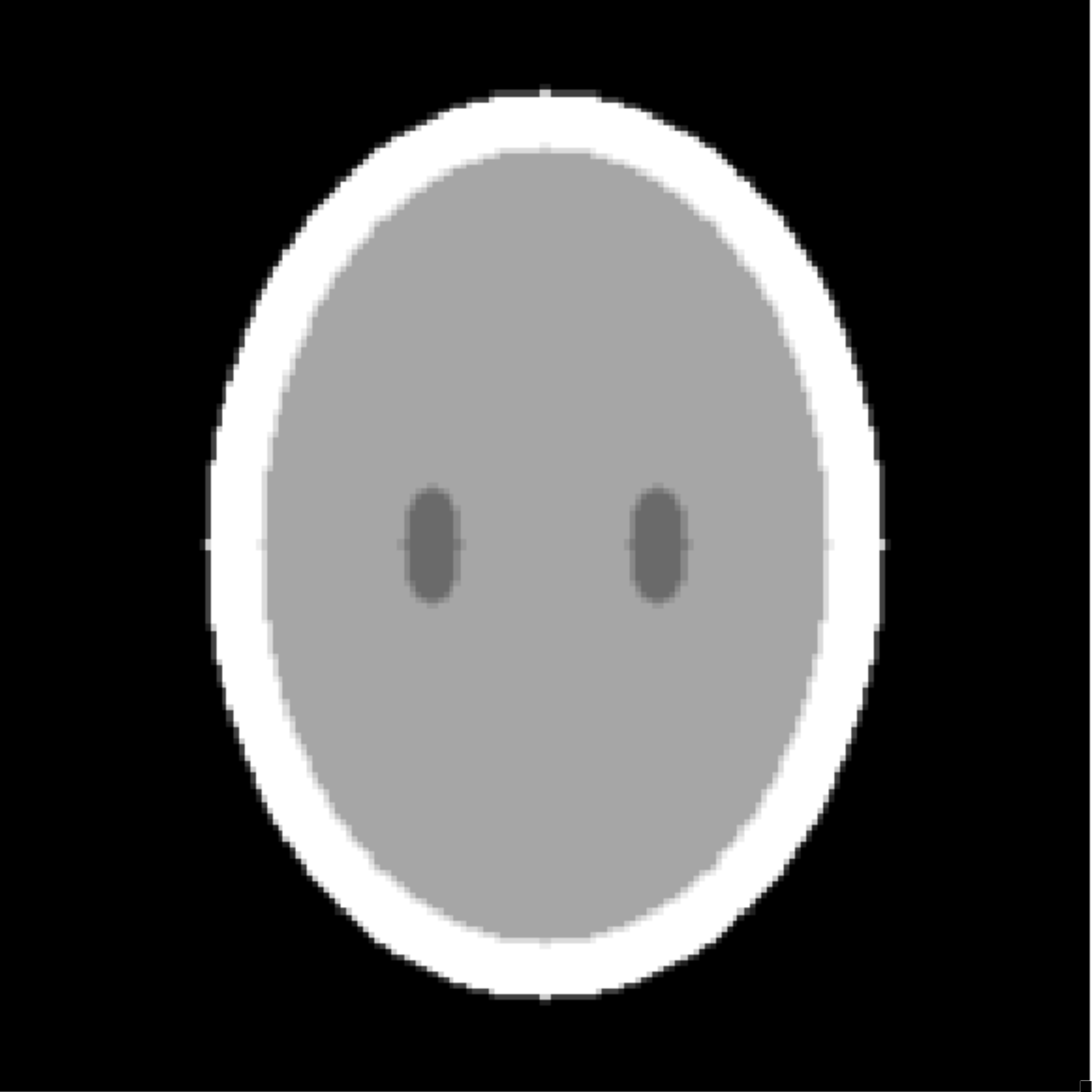}}
\end{minipage}
\caption{Digital head phantom used to generate simulated data in this work.}
\label{fig:phantom}
\end{figure}

The specific digital phantom used in this work is shown in Figure~\ref{fig:phantom}; the phantom had an isotropic voxel size of 1~mm$^{3}$ and was comprised of an outer elliptical region representing skull bone enclosing brain and two inner elliptical sections representing fluid-filled ventricles. These regions were assigned realistic RSP values: 1.6 for bone, 1.04 for brain, and 0.9 for ventricles.

Two simulated data sets were produced to assess the hull-detection algorithms presented in this paper, each with a total of 11,796,480 proton histories.  The first simulation in this study generated a uniform beam of 200~MeV protons.  Each proton in the beam was randomly distributed about the central beam axis to simulate a realistic proton cone-beam.  The simulator also generated bivariate normal random variables for exit angle and displacement with respect to the entry angle and displacement to simulate the effect of multiple Coulomb scattering inside the object.  The intersection lengths for the voxels that lay on a proton path were taken to be 1~mm so that the entry in every row of the system matrix was either one or zero. The noiseless WEPL generated for each proton was the sum of the RSP of each intersected voxel. Straight-line paths were assumed between the entry and exit points of the phantom.

The data from the first simulation was duplicated with noisy WEPL values. These were created by first converting a noiseless WEPL value into exit energy, generating a normally distributed noisy energy value with a standard deviation described by Tschalar's energy straggling theory \cite{Tschalar68}, and converting that noisy energy value back to a final WEPL value. The conversion of exit energy to WEPL and vice versa was based on ICRU Report 49~\cite{ICRU93}.
\subsection{Scanner Data}
\begin{figure}[h!]
\hspace{0.5cm}
  \begin{minipage}[b]{0.45\linewidth}
  \centering
  \centerline{\includegraphics[height=4.0cm]{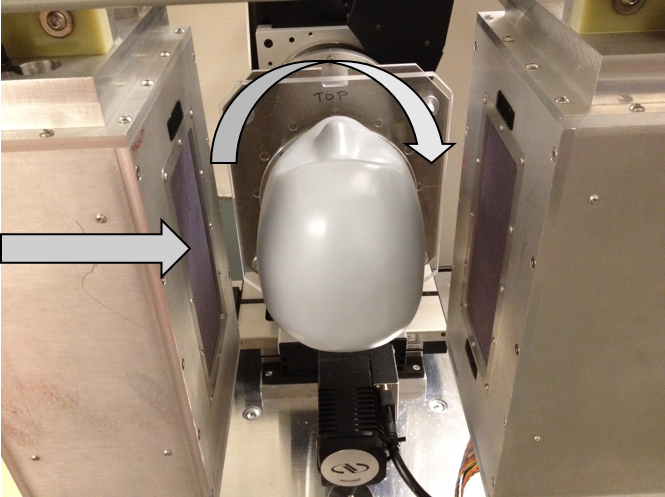}}
  \centerline{(a)}\medskip
\end{minipage}
\begin{minipage}[b]{0.45\linewidth}
  \centering
  \centerline{\includegraphics[height=4.0cm]{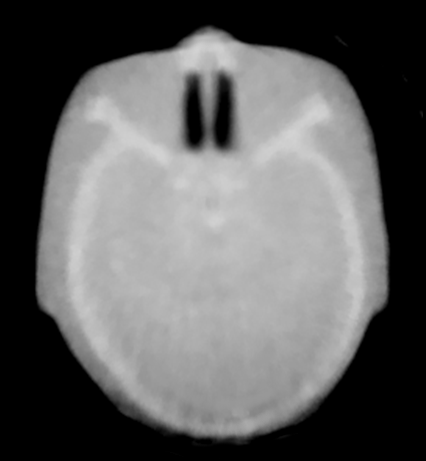}}
  \centerline{(b)}\medskip
\end{minipage}
\caption{(a) A pediatric head phantom being rotated (curved arrow) with respect to the fixed horizontal beam (straight arrow) on the Phase I prototype pCT scanner at Loma Linda University Medical Center. (b) pCT reconstruction of a representative slice of the pediatric head phantom.}
\label{fig:rotation}
\end{figure}

An experimental data set was used in order to assess the performance of each algorithm with realistic data.  This data set contained 50,897,953 proton histories obtained from a scan of an anthropomorphic pediatric head phantom (Model 715-HN, CIRS\footnote{http://www.cirsinc.com/products/all/36/pediatric-anthropomorphic-training-phantoms/?details=specs}) on the Phase I prototype pCT detector system shown in Figure~\ref{fig:rotation}.  The scan was obtained using a proton cone-beam of approximately 200~MeV generated by the medical proton accelerator at Loma Linda University Medical Center.  The number of proton histories corresponds to a complete data set, i.e., without the removal of unsuitable proton histories.  The pediatric head phantom was rotated a full 360$^{\circ}$ with respect to the fixed horizontal beam and pCT detector system in 4$^{\circ}$ increments.

Some of the proton histories may not be suitable for pCT image reconstruction, including hull-detection. One source of unsuitable proton histories in our present data sets was pile-up due to protons arriving at the energy detector (calorimeter) too closely in time, thus causing the residual energy from the first proton to be added to the energy measurement of the second proton. Other unsuitable proton histories had should be excluded from pCT image reconstruction because they underwent atypical physical interactions, including elastic large angle scattering and inelastic nuclear interactions.

Unsuitable data resulting from pile-up and atypical physical events can be removed by grouping (binning) histories into intervals with similar angle and similar horizontal and vertical displacement relative to the center of the reconstruction volume. Histories whose WEPL, relative horizontal angle, or relative vertical angle are beyond three standard deviations from the mean of each bin are then removed from the data set (data cuts).
\section{Hull-Detection Algorithms}
\label{sec:Methods}
\label{ssec:Hull Detection Algorithms}
For an object $X\subset\mathbb{N}^3$ with hull $H\subseteq V\subset\mathbb{N}^3$, a hull-detection algorithm seeks an approximate hull, $H'\subseteq V$, such that $H\subseteq H'$ and the cardinality, $\lvert H' \setminus H \rvert$, of the set difference, $H'\setminus H$, is as small as possible.  In essence, we seek to produce an approximation of the hull which contains every voxel of the object while minimizing the number of voxels included from outside the object.

Three algorithms were tested in this work and compared to filtered backprojection (FBP): silhouette/space carving (SC), modified silhouette/space carving (MSC), and space modeling (SM).  FBP is capable of performing a full image reconstruction; however, protons follow curved paths due to multiple Coulomb scattering, which does not fit the reconstruction framework of FBP.  Here, FBP was used to detect an approximate hull and to generate the starting data set for the iterative reconstruction algorithm, as done in previous pCT reconstruction work~\cite{PenfoldDis,PSCR10}.  In previous work on hull-detection algorithms for pCT reconstruction, we had compared SC to FBP with respect to computation time and quality \cite{SWSS12}, but that comparison did not include a voxel-by-voxel comparison between the true and the detected object hull.
\subsection{Filtered Backprojection (FBP)}
\label{ssec:FBP}
FBP is a well-known algorithm, first introduced for reconstruction of CT data by Ramachandran and Lakshminarayanan \cite{RL71}. In this work, we used the Feldkamp Davis Kress (FDK) algorithm \cite{fdk}, a cone-beam variant of the FBP algorithm, assuming that all proton paths through the object were straight lines and follow a cone-beam geometry. The FDK algorithm was performed with 4$^{\circ}$ angular bin spacing, a 1~mm lateral bin size, and a 5~mm vertical bin size. Each slice of the reconstruction volume was defined to be 200~mm~$\times$~200~mm and 3~mm thick. With this thickness and a reconstruction volume height of 9.6~cm, a total of 32 slices were produced. A Shepp-Logan filter~\cite{SL74} was used prior to backprojection. The resulting image was then thresholded to generate the approximate hull. Any voxel with RSP~$\geq 0.6$ was assumed to belong to the object and was assigned an RSP value of one. Voxels with RSP values below this threshold were assigned an RSP value of zero.

In this work, FBP was performed for hull-detection using the proton histories that remained after data cuts were performed to remove unsuitable proton histories. The data cuts were performed on the proton histories after binning them into the intervals described above.
\subsection{Silhouette/Space Carving (SC)}
\label{sssec:SC}
SC is an algorithm used to generate an approximation of the object's hull in a similar way that sculptures are chiseled from a solid block of material \cite{Niem94,Niem97,KS99}. If a proton does not pass through the object, it will not experience significant energy loss or scattering.  Therefore, individual or bin-averaged energy measurements (or, alternatively, converted WEPL values) and angular deviations can be used, in principle, to identify which protons missed the object entirely.  Protons that missed the object are identified by placing a cutoff value on energy measurements (or WEPL values) and angular deviations such that, if the measured values fall below the cutoff values, then the proton or all protons associated with a proton bin are assumed to have missed the object.

Initially, the object is assumed to encompass the entire reconstruction volume and each voxel is assigned an RSP value of one.  If a proton is identified as having missed the object, the voxels along its path, approximated by a straight line, are carved from the reconstruction volume by assigning them an RSP value of zero.  Note that a straight line is an accurate approximation of the path due to the insignificant amount of scattering.  The voxels that are not carved from the reconstruction volume and, thus, retain an RSP value of one, are then assumed to belong to the object hull.

SC (see Algorithm~\ref{alg:sc} below) seeks an approximation, $H_1$, of the hull, $H$, by identifying the protons that missed the object based on analysis of WEPL values and then carving the voxels along each of their paths from the reconstruction volume.  In other words, for each projection angle in the scan, projecting the paths of protons that miss the object through the reconstruction volume produces a silhouette of the object where no proton passed through.  Excluding the voxels outside these backprojected silhouettes then yields an approximate object hull.

Notationally, $p_i$ refers to the $i^{th}$ proton or proton bin, $\Delta E(p_i)$ refers to the energy loss of the $i^{th}$ proton or the mean energy loss of protons in bin $i$, respectively, and $\Delta\angle(p_i)$ is the change in angle of the $i^{th}$ proton or the mean change in angle of protons in bin $i$, respectively.
Let $I$ be the set of indices of all the protons or proton bins.
Let $E_L$ be the user-defined cutoff value on the energy lost in air and $\theta_L$ be the user-defined cutoff value on the angular change of protons in air.  If $\Delta E(p_i)$ and $\Delta\angle(p_i)$ fall below their associated cutoff values, then the proton or protons assigned to the proton bin, $p_i$, are assumed to have missed the object.  We then define the set, $I_L$, of indices of these protons or proton bins as
\begin{align}
    I_L=\left\{i\in I \mid \Delta E(p_i)<E_L, \, \Delta\angle(p_i)<\theta_L\right\}.
\end{align}
Let $V$ be the set of all voxels in the reconstruction space.
Let $L_i$ be the line that connects the entry and exit points of the proton or protons assigned to the proton bin with index $i$, for $i\in I_L$.
Given a distance measure $d(\cdot,\cdot)$ and a minimum distance $d_0$, we define the set, $A_i$, of voxels along the path, $L_i$, as
\begin{align}
   A_i = \left\{ v\in V \mid d(L_i,v)\leq d_0\right\}
\end{align}
and the approximate hull is then given by
\begin{align}
    H_1 = V \setminus \cup_{i\in I_L}A_i.
\end{align}
\begin{algorithm}
\caption{Silhouette/Space Carving (SC)}\label{alg:sc}
\begin{algorithmic}[1]
\State {$I_L \leftarrow \emptyset$}
\ForAll{$i \in I$}
    \If {$\Delta E(p_i)<E_L$ \textbf{and} $\Delta\angle(p_i)<\theta_L$}
        \State {$I_L \leftarrow \{I_L,i\}$}
    \EndIf
\EndFor
\ForAll{$v \in V$}
    \State {$H_1(v) \leftarrow 1$}
\EndFor
\ForAll{$i \in I_L$}
        \ForAll{$v \in V$}
            \If {$d(L_i,v)\leq d_0$}
                \State {$H_1(v) \leftarrow 0$}
            \EndIf
        \EndFor
\EndFor
\end{algorithmic}
\end{algorithm}

In this work, the implementation of the SC algorithm used the proton histories that remained after data cuts were performed to remove unsuitable proton histories. The bin size for the data cuts was the same as that used for the FDK algorithm. The same bins were also used to define the paths, $L_i$, in the SC algorithm.  A cutoff value on the mean WEPL of a bin, rather than the energy loss, was used to identify protons that missed the object; if the mean WEPL of a bin was less than 1.0~mm, the protons in that bin were assumed to have missed the object and the voxels along their path, approximated by a straight line using the angle and displacements associated with that bin, were carved from the reconstruction volume.  No angular cutoff was used in this implementation of the SC algorithm.

To avoid excluding valid portions of the object from the detected hull due to unsuitable proton histories evading data cuts and protons skimming the surface of the object, a simple $5\times 5$ averaging filter was applied to the image of the approximate hull.  The approximate object hull was then formed from the filtered image by assigning voxels a value of one if their value exceeded a threshold of $0.4$ (a tunable parameter) and assigned a value of zero otherwise.
\subsection{Modified Silhouette/Space Carving (MSC)}
\label{sssec:MSC}
SC does not record the number of times, $N$, a voxel was determined to lie outside the boundary of the object; thus, voxels belonging to the object can mistakenly be excluded from the detected hull due to the presence of unsuitable data.  MSC is a new hull-detection algorithm proposed here which uses the number of times, $N$, a voxel was identified as lying outside the object to determine which voxels should be excluded from the approximate hull.  Since unsuitable proton histories make up a relatively small percentage of the total proton histories, they will have minimal effect on $N$ for a particular voxel.  Therefore, by considering $N$, MSC can theoretically avoid mistakenly excluding voxels belonging to the object from the detected object hull.

MSC (see Algorithm~\ref{alg:msc} below) seeks to robustly generate an approximation, $H_2$, of the hull, $H$, by backprojecting the silhouette and counting the number of times, $N$, a voxel is marked outside the silhouette. Note that MSC uses only the proton histories that did not pass through the object.  We then define the set, $C(v)$, of indices of the paths, $L_i$, that passed through voxel $v$ as
\begin{align}
    C(v) =\left\{i\in I_L \mid v\in A_i\right\}
\end{align}
and the set, $B(v)$, of neighboring voxels, $w$, of voxel $v$ as
\begin{align}
    B(v) = \left\{ w\in V \mid d(v, w) \leq 1\right\}.
\end{align}
Defining the cardinality of $C(v)$ as $N(v) =\lvert C(v)\lvert$, the cardinality of $C(w)$ as $N(w) =\lvert C(w)\lvert$, and given a minimum cardinality difference threshold $N_t$, the approximate hull is then defined as
\begin{align}
    H_2 = \left\{v\in V \,\middle |\, \max_{w\in B(v)}N(v)-N(w) < N_t\right\}.
\end{align}
\begin{algorithm}
\caption{Modified Silhouette/Space Carving (MSC)}\label{alg:msc}
\begin{algorithmic}[1]
\State {$I_L \leftarrow \emptyset$}
\ForAll{$i \in I$}
    \If {$\Delta E(p_i)<E_L$ \textbf{and} $\Delta\angle(p_i)<\theta_L$}
        \State {$I_L \leftarrow \{I_L,i\}$}
    \EndIf
\EndFor
\ForAll{$v \in V$}
    \State {$H_2(v) \leftarrow 1$}
    \State {$N(v) \leftarrow 0$}
\EndFor
\ForAll{$i \in I_L$}
    \ForAll{$v \in V$}
        \If {$d(L_i,v)\leq d_0$}
            \State {$N(v) \leftarrow N(v) + 1$}
        \EndIf
    \EndFor
\EndFor
\ForAll{$v \in V$}
    \ForAll{$w \in B(v)$}
        \If {$N(v) - N(w) \geq N_t$}
            \State {$H_2(v) \leftarrow 0$}
        \EndIf
    \EndFor
\EndFor
\end{algorithmic}
\end{algorithm}

In this work, the implementation of the MSC algorithm did not include performing data cuts to remove unsuitable proton histories.  Decisions on whether protons missed the object were based on analysis of individual WEPL values rather than bin averages.  Thus, proton histories were not binned in this case. Proton histories whose WEPL values were less than 1.0~mm were assumed to have missed the object, which is the same WEPL cutoff value used for SC. A threshold $N_t=50$ was used here as this was found to be insensitive to the varying number of proton histories in each data set.  Note that a cardinality threshold proved to be an inadequate basis for identifying voxels outside the object hull, as an appropriate cardinality threshold varied between slices and data sets.  However, the cardinality difference, $N(v)-N(w)$, between neighboring voxels proved to be a more robust measure.
\subsection{Space Modeling (SM)}
\label{sssec:SM}
SM is a new hull-detection algorithm proposed here that only uses protons passing through the object to generate an approximation of an object hull.  If a proton passes through the object, it is likely going to experience energy loss and/or angular deviations, both of which tend to increase as the amount of material it passes through increases.  Similar to SC, protons can then be identified as having passed through the object based on energy measurements (or converted WEPL values) and angular deviations.  Protons that passed through the object are identified by placing a cutoff value on energy measurements (or WEPL values) and angular deviations such that, if the measured values exceed the cutoff values, then the proton is assumed to have passed through the object.

Protons identified as having passed through the object are assumed to have followed a straight line path and the voxels intersected by this path are determined.  The number of times, $M$, a voxel is intersected by the path of a proton that passed through the object is recorded for each voxel.  Similar to MSC, recording $M$ helps avoid the effects of misidentifying protons as having passed through the object due to unsuitable proton histories.  However, this is a necessary aspect of SM because protons that pass through the object pass through voxels that are not part of the object hull.  Thus, $M$ helps discern between voxels belonging to the object hull and those that lie outside its boundary.

Based on our experience, $M$ drops more sharply at the boundary of the object than in any other location.  The edge with the largest gradient in $M$ is located automatically in each slice and the largest value of $M$ on this edge is used to set the threshold, $M_t$, for that slice.  Any voxel in that slice with $M\geq M_t$ is then assumed to be part of the object.  This process is then repeated for each slice to generate the approximate hull.

Similar to MSC, SM (see Algorithm~\ref{alg:sm} below) seeks to robustly generate an approximate hull, $H_3$, by backprojecting the silhouette and counting the number of times, $M$, a voxel is identified as part of the object.  Let $E_H$ and $\theta_H$ be the user-defined cutoff values on the energy loss and angular change, respectively, used to determine if a proton passed through an object.  If $\Delta E(p_i)$ or $\Delta\angle(p_i)$ exceed these cutoff values, then the proton or protons assigned to the proton bin, $p_i$, are assumed to have passed through the object.  We then define the set, $I_H$, of indices of these protons or proton bins as
\begin{align}
    I_H=\left\{i\in I \mid \Delta E(p_i)>E_H\right\}.
\end{align}
Let $V$ be the set of all voxels in the reconstruction space.
Let $L_i$ be the line that connects the entry and exit points of the proton or protons assigned to the proton bin with index $i$, for $i\in I_H$.
Given a distance measure $d(\cdot,\cdot)$ and a minimum distance $d_0$, we define the set, $A_i$, of voxels along the path, $L_i$, as
\begin{align}
    A_i = \left\{ v\in V \mid d(L_i,v)\leq d_0\right\},
\end{align}
and the set, $C(v)$, of indices of the paths, $L_i$, that passed through voxel $v$ as
\begin{align}
    C(v) =\left\{i\in I_H \mid v\in A_i\right\}.
\end{align}
Defining the cardinality of $C(v)$ as $M(v) =\lvert C(v)\lvert$, the cardinality of $C(w)$ as $M(w) =\lvert C(w)\lvert$, and given a minimum cardinality threshold $M_t$, the approximate hull is then defined as
\begin{align}
    H_3 = \left\{v \mid M(v) >M_t\right\}.
\end{align}
\begin{algorithm}
\caption{Space Modelling (SM)}\label{alg:sm}
\begin{algorithmic}[1]
\State {$I_H \leftarrow \emptyset$}
\ForAll{$i \in I$}
    \If {$\Delta E(p_i)>E_H$ \textbf{or} $\Delta\angle(p_i) > \theta_H$}
        \State {$I_H \leftarrow \{I_H,i\}$}
    \EndIf
\EndFor
\ForAll{$v \in V$}
    \State {$H_3(v) \leftarrow 0$}
    \State {$M(v) \leftarrow 0$}
\EndFor
\ForAll{$i \in I_H$}
    \ForAll{$v \in V$}
        \If {$d(L_i,v)\leq d_0$}
            \State {$M(v) \leftarrow M(v) + 1$}
        \EndIf
    \EndFor
\EndFor
\State {$MaxSlope\leftarrow 0$}
\State {$index\leftarrow 0$}
\ForAll{$v \in V$}
    \ForAll{$w \in B(v)$}
        \If {$M(v)- M(w) \geq MaxSlope$}
            \State {$MaxSlope\leftarrow M(v)- M(w)$}
            \State {$index\leftarrow v$}
        \EndIf
    \EndFor
\EndFor
\State {$M_t \leftarrow  M(index)$}
\ForAll{$v \in V$}
    \If {$M(v) > M_t$}
        \State {$H_3(v) \leftarrow 1$}
    \EndIf
\EndFor
\end{algorithmic}
\end{algorithm}

In this work, the implementation of the SM algorithm, like the MSC algorithm, did not include data cuts to remove outliers and individual (rather than bin-averaged) WEPL values were used to determine if a proton passed through the object.  A WEPL cutoff value of 5.0~mm and no angular cutoff was used in this implementation.  We picked the minimum cardinality threshold, $M_t$, for each slice using a modified version of the Canny edge detection algorithm~\cite{JC86}.  Note that the neighborhood comparison method used in MSC did not work well with SM.
\section{Results}
\label{sec:Results}
\subsection{Simulated Data Results}
\begin{figure}[h!]
\begin{minipage}[t]{.19\linewidth}
  \centering
  \centerline{\includegraphics[width=2.5cm]{NeoSimZdim.pdf}}
  \centerline{(a) Phantom}\medskip
\end{minipage}
\hfill
\begin{minipage}[t]{.19\linewidth}
  \centering
 \centerline{\includegraphics[width=2.5cm]{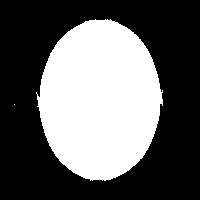}}
  \centerline{(b) FBP}
\end{minipage}
\hfill
\begin{minipage}[t]{0.19\linewidth}
  \centering
  \centerline{\includegraphics[width=2.5cm]{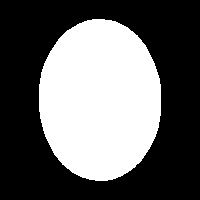}}
  \centerline{(c) SC}
\end{minipage}
\hfill
\begin{minipage}[t]{0.19\linewidth}
  \centering
  \centerline{\includegraphics[width=2.5cm]{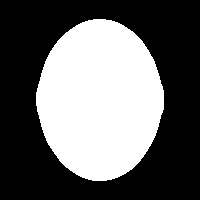}}
  \centerline{(d) MSC}
\end{minipage}
\hfill
\begin{minipage}[t]{0.19\linewidth}
  \centering
  \centerline{\includegraphics[width=2.5cm]{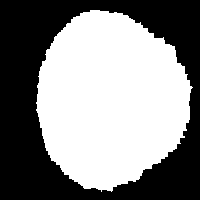}}
  \centerline{(e) SM}
\end{minipage}
\caption{ (a) Original digital head phantom; (b)-(e) object hull approximations generated by the various hull-detection algorithms for the noiseless simulated data set.}
\label{fig:noiseless}
\end{figure}

\begin{figure}
\begin{minipage}[t]{.19\linewidth}
  \centering
  \centerline{\includegraphics[width=2.5cm]{NeoSimZdim.pdf}}
  \centerline{(a) Phantom}\medskip
\end{minipage}
\hfill
\begin{minipage}[t]{.19\linewidth}
  \centering
 \centerline{\includegraphics[width=2.5cm]{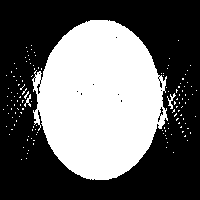}}
  \centerline{(b) FBP}
\end{minipage}
\hfill
\begin{minipage}[t]{0.19\linewidth}
  \centering
  \centerline{\includegraphics[width=2.5cm]{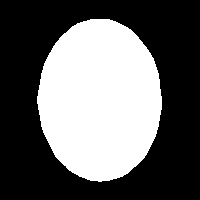}}
  \centerline{(c) SC}
\end{minipage}
\hfill
\begin{minipage}[t]{0.19\linewidth}
  \centering
  \centerline{\includegraphics[width=2.5cm]{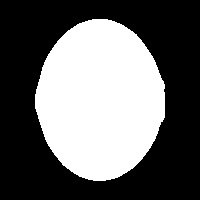}}
  \centerline{(d) MSC}
\end{minipage}
\hfill
\begin{minipage}[t]{0.19\linewidth}
  \centering
  \centerline{\includegraphics[width=2.5cm]{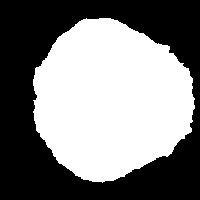}}
  \centerline{(e) SM}
\end{minipage}
\caption{(a) Original digital head phantom; (b)-(e) object hull approximations generated by the various hull-detection algorithms for the noisy simulated data set.}
\label{fig:noisy}
\end{figure}

Figure~\ref{fig:noiseless} shows a visual representation of the hull approximations generated by each algorithm using the noiseless simulated data set for a single slice of a NEO, simulating a head.  The dimensions of all images are 200~voxels $\times$ 200~voxels and represent an area of 200~mm $\times$ 200~mm.  Table~\ref{tab:noiseless} summarizes the computation times and the number of missing and extra voxels, respectively, resulting from a voxel-by-voxel comparison between the known digital head phantom slice and each approximate hull resulting from analysis of the noiseless simulated data set.  Note that the original phantom slice contained 15,336 voxels.  The results for the noisy simulated data set are shown in Figure~\ref{fig:noisy} and in Table~\ref{tab:noisy}, respectively.
\begin{table}[h!]
\centering
\caption{Comparison of hull-detection algorithms for noiseless data set}
\label{tab:noiseless}
\begin{tabular}{c c c c c}
&FBP&SC&MSC&SM\\ \hline\hline
Computation Time &16.70~s&$<$0.10~s&5.95~s&5.52~s\\
Missing Voxels&50&0&0&0\\
Extra Voxels&116&345&488&5802\\\hline\hline
\end{tabular}
\end{table}
\begin{table}[h!]
\centering
\caption{Comparison of hull-detection algorithms for noisy data set}
\label{tab:noisy}
\begin{tabular}{c c c c c}
&FBP&SC&MSC&SM\\ \hline\hline
Computation Time &16.72~s&$<$0.10~s&6.14~s&5.86~s\\
Missing Voxels&88&0&0&0\\
Extra Voxels&831&461&716&4563\\\hline\hline
\end{tabular}
\end{table}

FBP hull-detection was the only algorithm that led to missing voxels for both the noiseless and the noisy simulated data sets. In addition, FBP produced falsely detected hull points outside the phantom hull, which led to visible artifacts in Figures~\ref{fig:noiseless}(b) and \ref{fig:noisy}(b), particularly for the hull approximation from the noisy simulated data set.  None of the hull approximations generated by the other hull-detection algorithms (SC, MSC, and SM) had missing voxels; however, the algorithms differed in the number of extra voxels that were falsely identified as belonging to the object's hull. Here, SC performed best and SM performed worst, with MSC showing an intermediate result.  Without noise, the approximate hull generated by FBP had the smallest number of extra voxels, whereas for the noisy data set, it had the second largest number after SM.

When comparing computation times (Tables~\ref{tab:noiseless} and \ref{tab:noisy}), SC was by far the fastest of the hull-detection algorithms, two orders of magnitude faster than the other algorithms.  MSC and SM performed similarly and FBP was consistently the slowest hull-detection algorithm.  This is understandable because FBP performs a full image reconstruction that then needs to be thresholded to approximate the object hull.
\subsection{Experimental Data Results}
\begin{figure}[h!]
\begin{minipage}[h!]{.19\linewidth}
  \centering
 \centerline{\includegraphics[width=2.5cm]{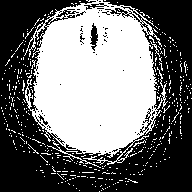}}
  \centerline{(a) FBP}
\end{minipage}
\hfill
\begin{minipage}[h!]{0.19\linewidth}
  \centering
  \centerline{\includegraphics[width=2.5cm]{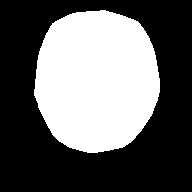}}
  \centerline{(b) SC}
\end{minipage}
\hfill
\begin{minipage}[h!]{0.19\linewidth}
  \centering
  \centerline{\includegraphics[width=2.5cm]{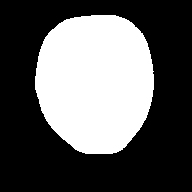}}
  \centerline{(c) MSC}
\end{minipage}
\hfill
\begin{minipage}[h!]{0.19\linewidth}
  \centering
  \centerline{\includegraphics[width=2.5cm]{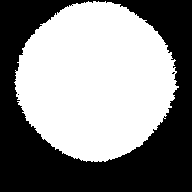}}
  \centerline{(d) SM}
\end{minipage}
\caption{(a)-(d) Object hull approximations generated by the various hull-detection algorithms using the experimental data set from the scan of the pediatric head phantom.}
\label{fig:pedhead}
\end{figure}

Figure~\ref{fig:pedhead} shows the hull approximations of a single representative slice of the pediatric head phantom generated by each algorithm using the experimental data set from the scan of the phantom.  The dimensions of all images are 192~voxels $\times$ 192~voxels and represent an area of 200~mm $\times$ 200~mm.

Compared to the other hull-detection algorithms (SC, MSC, SM), the approximate hull generated by FBP (Figure~\ref{fig:pedhead}(a)) contained the largest number of artifacts, i.e., streaks outside the object and missing voxels inside the object. The FBP hull-detection algorithm also recognized the nasal cavity inside the phantom as not belonging to the hull because it contained RSP values below the threshold of 0.6. Notice that FBP falsely identified three nasal passages, whereas the real phantom only contains two passages separated by the nasal septum.

The hull approximations generated by SC (Figure~\ref{fig:pedhead}(b)) and MSC (Figure~\ref{fig:pedhead}(c)) are free of both interior and exterior artifacts, and both approximations appear to generally match the outline of the real phantom, although a direct comparison with the true object hull was not possible in this case. On the other hand, the approximate hull generated by SM (Figure~\ref{fig:pedhead}(d)) was clearly larger than the other two hull approximations and its shape indicates that it contained a substantial number of extra voxels not belonging to the true hull, matching the results from the simulated data sets.
\section{Discussion}
\label{sec:Summary}
The work presented here is an extension of previous work on detecting the hull of objects as part of an iterative pCT reconstruction algorithm~\cite{SWSS12}. In the previous results, we had considerable artifacts related to unsuitable proton histories. In this work, we identified possible solutions to avoid these artifacts.  For SC, we removed unsuitable histories from the data sets by performing data cuts.  In addition, we binned the data into intervals, only using the mean value of each bin in the hull-detection algorithms, and introduced a blurring filter to fill in voxels mistakenly excluded from the approximate hull.

We also introduced two novel hull-detection algorithms (MSC and SM) for which binning and removal of unsuitable data was not used for hull-detection.  For the hull-detection of the NEO slice with simulated data sets, this led to accurate identification of voxels belonging to the object.  Whether this is also the case for experimental data, with more significant outliers in the data sets, remains to be determined.

In the present work, MSC and SM did not require binning the data for adequate hull detection. Therefore, it may be possible to do the carving of each path from the approximate hull independently in an online mode during data acquisition. Thus, although MSC and SM are substantially more computationally expensive than SC, they could be computationally ecient if they can be executed during the scan. The results of MSC and SM would then become available almost immediately after the scan, whereas the SC algorithm would need to wait until all data have been acquired.

Considering the accuracy of the object hull approximations, SC was clearly the most viable hull-detection algorithm. Future work will investigate whether by adjusting threshold parameters or by providing better data selection, MSC and SM can be further improved to perform equally well, or even better, than SC. Based on the present work, hull-detection using an FBP algorithm does not produce satisfactory results. FBP is also not competitive in terms of computational efficiency.
\section{Conclusion}
This work has investigated the suitability of two existing and two new hull-detection algorithms for pCT reconstruction. The results obtained with SC, MSC, and SM are promising and represent a significant step toward an effective and robust hull-detection algorithm. SC performed best, but MSC and SM could be further improved. FBP was not adequate for efficient and accurate hull-detection.
\bibliographystyle{amsalpha}
\bibliography{pct}

\newcommand{\etalchar}[1]{$^{#1}$}
\providecommand{\bysame}{\leavevmode\hbox to3em{\hrulefill}\thinspace}
\providecommand{\MR}{\relax\ifhmode\unskip\space\fi MR }
\providecommand{\MRhref}[2]{%
  \href{http://www.ams.org/mathscinet-getitem?mr=#1}{#2}
}
\providecommand{\href}[2]{#2}
\begin{thebibliography}{CCC{\etalchar{+}}07}

\bibitem[ar68]{Tschalar68}
C.~Tschal\" ar, \emph{Straggling distributions of extremely large energy
  losses}, Nucl. Instrum. Methods \textbf{61} (1968), 141--156.

\bibitem[BBF{\etalchar{+}}07]{Fel07}
M.~Bruzziand, N.~Blumenkrantz, J.~Feldt, J.~Heimann, H.F.-W. Sadrozinski,
  A.~Seiden, D.C. Williams, V.~Bashkirov, R.~Schulte, D.~Menichelli,
  M.~Scaringella, G.G.P. Cirrone, G.~Cuttone, N.~Randazzo, V.~Sipala, and D.~Lo
  Presti, \emph{Prototype tracking studies for proton {CT}}, IEEE Trans. Nucl.
  Sci. \textbf{54} (2007), 140--145.

\bibitem[Can86]{JC86}
J.~Canny, \emph{A computational approach to edge detection}, IEEE Trans.
Pattern Anal. Mach. Intell. \textbf{8} (1986), 679--698.

\bibitem[CCC{\etalchar{+}}07]{Cut07}
G.~Cuttone, G.A.P. Cirrone, G.~Candiano, F.~Di Rosa, G.~Russo, N.~Randazzo,
  V.~Sipala, S.~Lo Nigro, D.~Lo Presti, J.~Feldt, J.~Heimann, H.F.-W.
  Sadrozinski, A.~Seiden, D.C. Williams, V.~Bashkirov, R.~Schulte, M.~Bruzzi,
  and D.~Menichelli, \emph{Monte carlo studies of a proton computed tomography
  system}, IEEE Trans. Nucl. Sci. \textbf{54} (2007), 1487--1491.

\bibitem[FDK84]{fdk}
L.A. Feldkamp, L.C. Davis, and J.W. Kress, \emph{Practical cone-beam
  algorithms}, J. Opt. Soc. Am. \textbf{A1} (1984), 612--619.

\bibitem[{Int}93]{ICRU93}
{International Commission on Radiation Units and Measurements}, \emph{Stopping
  powers and ranges for protons and alpha particles}, ICRU Report \textbf{49}
  (1993).

\bibitem[KS99]{KS99}
K.N. Kutulakos and S.M. Seitz, \emph{A theory of shape by space carving},
  Proc. Seventh International Conference on Computer Vision (ICCV), 1999,
  307--314.

\bibitem[Nie94]{Niem94}
W.~Niem, \emph{Robust and fast modelling of 3{D} natural objects from multiple
  views}, SPIE Proceedings Image and Video Processing \textbf{2182} (1994),
  388--397.

\bibitem[Nie97]{Niem97}
\bysame, \emph{Error analysis for silhouette-based 3{D} shape estimation from
  multiple views}, Proc. Int. Workshop on Synthetic-Natural Hybrid Coding and
  Three-Dimensional Imaging, 1997, 143--146.

\bibitem[Pen10]{PenfoldDis}
S.N. Penfold, \emph{{Image Reconstruction and Monte Carlo Simulations in the
  Development of Proton Computed Tomography for Applications in Proton
  Radiation Therapy}}, Ph.D. thesis, University of Wollongong, Australia, 2010.

\bibitem[PSCR10]{PSCR10}
S.N. Penfold, R.W. Schulte, Y.~Censor, and A.B. Rosenfeld, \emph{Total
  variation superiorization schemes in proton computed tomography image
  reconstruction}, Med. Phys. \textbf{37} (2010), 5887--5895.

\bibitem[RL71]{RL71}
G.N. Ramanchandran and A.V. Lakshminarayanan, \emph{Three dimensional
  reconstructions from radiographs and electron micrographs: Application of
  convolution instead of {F}ourier transforms}, Proc. Natl. Acad. Sci. U.S.A.
  \textbf{68} (1971), 2236--2240.

\bibitem[SL74]{SL74}
L.~Shepp and B.~Logan, \emph{The {F}ourier reconstruction of a head section},
  IEEE Trans. Nucl. Sci. \textbf{NS-21} (1974), 21--43.

\bibitem[Smi09]{Smith09}
A.R. Smith, \emph{Vision 20/20: proton therapy}, Med. Phys. \textbf{36} (2009),
  556--568.

\bibitem[SPTS08]{SPTS08}
R.~Schulte, S.~Penfold, J.~Tafas, and K.E. Schubert, \emph{A maximum likelihood
  proton path formalism for application in proton computed tomography}, Med.
  Phys. \textbf{35} (2008), 4849--4856.

\bibitem[SWSS12]{SWSS12}
B.~Schultze, M.~Witt, K.E. Schubert, and R.~Schulte, \emph{Space carving and
  filtered back projection as preconditioners for proton computed tomography
  reconstruction}, Proceedings of the IEEE Nuclear Science Symposium Medical
  Imaging Conference, 2012, 4335--4340.

\bibitem[WSS{\etalchar{+}}12]{WSS12}
M.~Witt, B.~Schultze, R.~Schulte, K.E. Schubert, and E.~Gomez, \emph{A proton
  simulator for testing implementations of proton CT reconstruction algorithms
  on GPGPU clusters}, Proceedings of the IEEE Nuclear Science Symposium Medical
  Imaging Conference, 2012, 4329--4334.

\end{thebibliography}
\end{document}